\documentclass[conference]{IEEEtran}
\IEEEoverridecommandlockouts
\usepackage{amsmath,amssymb,amsfonts}
\usepackage{algorithmic}
\usepackage{graphicx}
\usepackage{textcomp}
\usepackage{xcolor}
\def\BibTeX{{\rm B\kern-.05em{\sc i\kern-.025em b}\kern-.08em
    T\kern-.1667em\lower.7ex\hbox{E}\kern-.125emX}}

\usepackage{multirow}
\usepackage{tabularray}
\usepackage[normalem]{ulem}

\definecolor{tab:darkviolet}{RGB}{148, 0, 211}
\definecolor{tab:bggrey}{RGB}{163, 163, 163}
\definecolor{tab:k}{RGB}{0, 0, 0}
\definecolor{tab:hotpink}{RGB}{255, 105, 180}
\definecolor{tab:black}{RGB}{148, 0, 211}
\definecolor{tab:r}{RGB}{255, 0, 0}
\definecolor{tab:b}{RGB}{0, 0, 255}
\definecolor{tab:olivedrab}{RGB}{107, 142, 35}
\definecolor{tab:deepskyblue}{RGB}{0, 191, 255}
\definecolor{tab:lime}{RGB}{0, 255, 0}

\newcommand \colorindicator[1]{%
	\begingroup%
	{\textcolor{#1}{\tiny{$\blacksquare \hspace{-.5mm} \blacksquare$}}}%
	\endgroup
}

\DeclareMathOperator{\atantwo}{atan2}

\makeatletter
\def\ps@IEEEtitlepagestyle{%
  \def\@oddhead{\mycopyrightnotice}%
  \def\@oddfoot{}
  \def\@evenhead{\@IEEEheaderstyle\thepage\hfil\leftmark\hbox{}}\relax
  \def\@evenfoot{}%
}
\def\mycopyrightnotice{%
  \begin{minipage}{\textwidth}
  \centering \scriptsize
  \textcolor{red}{Copyright~\copyright~2025 IEEE. Personal use of this material is permitted. Permission from IEEE must be obtained for all other uses, in any current or future media, including reprinting/republishing this material for advertising or promotional purposes, creating new collective works, for resale or redistribution to servers or lists, or reuse of any copyrighted component of this work in other works. 
  This is the accepted version of the manuscript. The final version is published in IEEE Access, DOI: {https://doi.org/10.1109/ACCESS.2025.3531667}.}
  \end{minipage}
}
\makeatother

\begin{document}

\title{Transformer-Based Model for Monocular Visual Odometry: A Video Understanding Approach

\thanks{
The work of André Françani was supported by the Coordination of Improvement of Higher Education Personnel (CAPES) under Grant 88887.008029/2024-00. The work of Marcos Maximo was supported in part by the National Research Council of Brazil (CNPq) under Grant 307525/2022-8.
}
}

\author{\IEEEauthorblockN{1\textsuperscript{st} André O. Françani}
\IEEEauthorblockA{\textit{Autonomous Computational Systems Lab (LAB-SCA)}\\
\textit{Computer Science Division} \\
\textit{Aeronautics Institute of Technology}\\
São José dos Campos, SP, Brazil \\
andre.francani@ga.ita.br}
\and
\IEEEauthorblockN{2\textsuperscript{nd} Marcos R. O. A. Maximo}
\IEEEauthorblockA{\textit{Autonomous Computational Systems Lab (LAB-SCA)} \\ 
\textit{Computer Science Division} \\
\textit{Aeronautics Institute of Technology}\\
São José dos Campos, SP, Brazil \\
mmaximo@ita.br}
}

\maketitle


\begin{abstract}
Estimating the camera's pose given images from a single camera is a traditional task in mobile robots and autonomous vehicles. This problem is called monocular visual odometry and often relies on geometric approaches that require considerable engineering effort for a specific scenario.
Deep learning methods have been shown to be generalizable after proper training and with a large amount of available data. Transformer-based architectures have dominated the state-of-the-art in natural language processing and computer vision tasks, such as image and video understanding. In this work, we deal with the monocular visual odometry as a video understanding task to estimate the 6 degrees of freedom of a camera's pose. We contribute by presenting the TSformer-VO model based on spatio-temporal self-attention mechanisms to extract features from clips and estimate the motions in an end-to-end manner. Our approach achieved competitive state-of-the-art performance compared with geometry-based and deep learning-based methods on the KITTI visual odometry dataset, outperforming the DeepVO implementation highly accepted in the visual odometry community. The code is publicly available at {https://github.com/aofrancani/TSformer-VO}.
\end{abstract}


\section{Introduction}
\label{sec:introduction}
Determining the location of a robot in an environment is a classical task for mobile robots and autonomous vehicles applications \cite{scaramuzza2011VO}. Visual odometry (VO) consists of estimating the camera's pose and motion given a sequence of frames, i.e. using visual sensors. In the monocular case, the system has a single camera to capture the images. Consequently, there is a lack of depth information when the three-dimensional (3D) objects are projected into the two-dimensional (2D) image space. 
Among the approaches to address the monocular VO, there are traditional geometry-based methods \cite{Geiger2011IV, ORBSLAM3_TRO}, deep learning-based methods that use end-to-end architectures \cite{wang2017deepvo, li2018undeepvo}, and hybrid methods with deep learning in certain modules of the geometry-based methods \cite{zhan2020visual, bruno2021lift, francani2022dense}.

Although traditional methods are robust and well-developed, they must be fine-tuned properly to achieve high performance, i.e. there is a requirement for considerable engineering effort for each specific application \cite{wang2017deepvo}. Furthermore, monocular VO systems suffer from scale ambiguity due to the lack of depth. The scale drift caused by the accumulation of scale errors during motion estimation over time can be reduced using additional sensors or some supplementary information \cite{scaramuzza2011VO}. 
On the opposing side, end-to-end approaches can estimate the 6 degrees of freedom (DoF) poses directly from a sequence of RGB images, that is, it does not need engineering effort to design the modules in the traditional VO pipeline \cite{wang2017deepvo}. However, its performance is strongly dependent on a large and diverse training dataset to make the model robust and reliable, which is typical of deep learning approaches.

In recent years, the Transformer architecture \cite{vaswani2017attention}, which is a deep learning method based on attention mechanisms, has been the basis of the state-of-the-art architectures in natural language processing (NLP) tasks, outperforming models based on recurrent neural networks (RNN) with long short-term memory cells (LSTM). After the success of transformer-based architectures such as BERT \cite{devlin2018bert} and GPT \cite{radford2018improving}, researchers started to apply the Transformer network to vision problems, reaching state-of-the-art performance on several image recognition benchmarks with the vision Transformer (ViT) model \cite{dosovitskiy2020image}. Furthermore, transformer-based architectures also achieved state-of-the-art results on video understanding tasks \cite{Arnab_2021_ICCV, gberta_2021_ICML}.

In this article, we propose an end-to-end architecture based on Transformer to estimate the 6-DoF pose in the context of visual odometry. We treat the visual odometry as a video understanding problem, where the 6-DoF camera's poses are estimated directly from a sequence of raw images. This approach is promising once visual odometry depends on spatio-temporal features, and transformer-based models have shown outstanding results on NLP tasks where sequential data is important.

We chose the TimeSformer \cite{gberta_2021_ICML} model for this work since it achieved state-of-the-art results in video action recognition tasks compared to previous works based on convolutional neural networks (CNN). In contrast to the original work, our idea is to use an MSE loss to perform regression instead of classification, together with a proper post-processing step to recover the pose estimation from overlapped-windowed data. We conducted experiments on the model hyperparameters and showed that our approach achieved competitive results on the KITTI odometry benchmark.
The main contributions of this work are summarized as follows:
\begin{itemize}
    \item We present a video understanding perspective for monocular visual odometry tasks, estimating the poses in a video clip all at once.
    \item We propose the TSformer-VO: an end-to-end supervised architecture based on spatio-temporal attention mechanisms that estimates the 6-DoF camera's pose.
    \item We show that our approach achieves competitive results on the KITTI odometry benchmark compared to previous state-of-the-art end-to-end methods based on deep learning.
\end{itemize}

The remaining of this paper is organized as follows. Section~\ref{sec:rel_work} presents related works in monocular visual odometry, showing methods based on geometry and deep learning, as well as presenting the video understanding task. Section~\ref{sec:method} introduces the proposed method and all required pre-processing and post-processing steps to estimate the 6-DoF camera's poses.
Section~\ref{sec:experiment} defines the experimental setups of our experiments, presenting the results and their discussions.
Finally, Section~\ref{sec:conclusion} establishes the overall conclusions of our work.

\section{Related work}\label{sec:rel_work}

In this Section, we introduce the monocular visual odometry methods based on geometry and deep learning, as well as the state-of-the-art techniques for the video understanding problem. 

\subsection{Monocular visual odometry}
Monocular visual odometry estimates incrementally the motion of an agent using a sequence of frames captured from a single camera attached to this agent \cite{scaramuzza2011VO}. The motion can be estimated from geometry constraints, or estimated from a deep learning approach without explicit engineering modeling of the scenario. 

\subsubsection{Geometry-based methods}
\begin{figure*}[!t]
\centering
\includegraphics[width=0.65\textwidth]{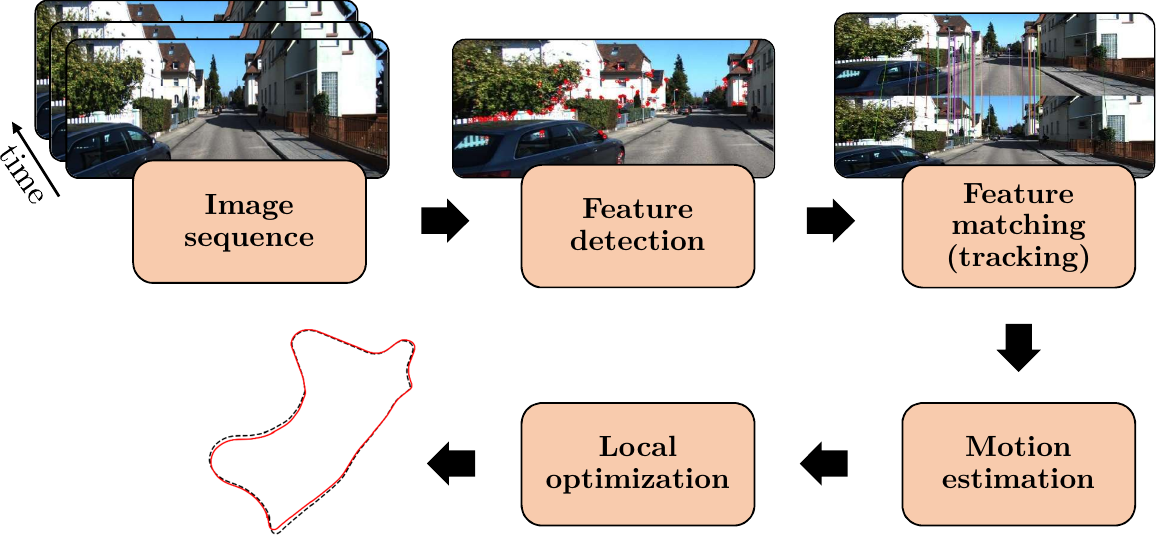}
\caption{Traditional pipeline for visual odometry. The scenario images are taken from the KITTI odometry dataset \cite{Geiger2012CVPR}.}
\label{fig:geometry-pipeline}
\end{figure*}

In general, geometry-based algorithms follow the pipeline depicted in Fig.~\ref{fig:geometry-pipeline}. Firstly, features or keypoints are detected in a frame, usually edges and corners. These features are matched and tracked to the subsequent frame using a similarity measurement. With the matched keypoints in successive frames, the essential matrix can be estimated using epipolar constraints. Then, the rotation matrix and translation vector are obtained by decomposing the essential matrix. Finally, the camera's poses can be refined through an offline local optimization, such as bundle adjustment \cite{scaramuzza2011VO}. 

LIBVISO2 \cite{Geiger2011IV} is a known implementation of the traditional pipeline for the monocular and stereo cases. The monocular system uses an 8-point algorithm to estimate the essential matrix, and it assumes that the camera has a fixed height over the ground for estimating the scale. 
Another widely recognized feature-based method is ORB-SLAM2 \cite{orbslam2}, which utilizes ORB keypoints for robust feature extraction. It integrates a mapping system that allows for place recognition, enabling loop closure and performing both local and global optimizations to refine the trajectory and map. Its most recent version, ORB-SLAM3 \cite{ORBSLAM3_TRO}, introduced improvements by extending the system to handle monocular, stereo, and RGB-D inputs within a unified framework. One of its key advancements is the inclusion of inertial information, allowing visual-inertial odometry.

\subsubsection{Deep learning methods}
Methods based on deep learning are alternatives for estimating the agent's pose from the data, imagery in the case of vision. These models can replace complex geometric modeling of the scene and feature extraction steps. However, in the visual odometry field, deep learning algorithms are more commonly used to replace one or more geometry-based odometry modules, which we refer to as hybrid approaches. 

Deep learning replacements in hybrid approaches can be applied to several steps, such as feature extraction \cite{duan2019deep}, feature matching \cite{sarlin2020superglue}, depth estimation \cite{ming2021deep}, and others \cite{arshad2021role}. DF-VO \cite{zhan2021df} is a successful example of a hybrid model, where the authors use classical geometric models to estimate the poses, while using deep learning models to estimate optical flow as feature matching. It also uses a monocular depth estimator based on deep learning to provide additional information in a scale recovery step. Another similar example is RAUM-VO \cite{cimarelli2022raum}, which uses pre-trained models to extract the features and perform the matching, along with the monocular depth estimation. The rotation is estimated by a geometric model, and it is adjusted by unsupervised training. Therefore, both methods are still hybrid approaches since they combine deep learning with geometry-based approaches.

Another common application of deep learning in the context of visual odometry is the fusion between imagery and multi-sensor data to improve the accuracy of pose estimation \cite{almalioglu2022selfvio}. In this case, researchers usually refer to visual-inertial odometry (VIO), where the image information is combined with an inertial measurement unit (IMU) to estimate the poses with high precision \cite{qin2022survey}. 

In addition to the mentioned approaches, some end-to-end models extract the 6-DoF pose purely by deep learning techniques directly from the data, without relying on any geometry-based modules or additional sensors.
End-to-end deep learning-based methods estimate the 6-DoF pose of a camera given a sequence of images. It has neither the feature detection and matching steps nor the geometry constraints. The main idea is to extract features from images like a common computer vision task, and use them for regression to estimate the camera's pose.

DeepVO \cite{wang2017deepvo} is a state-of-the-art end-to-end architecture that extracts features from an image pair (consecutive frames) using a CNN with pre-trained FlowNet weights \cite{dosovitskiy2015flownet}, followed by a RNN with LSTM cells to handle the temporal information of the estimated poses. DeepVO achieved competitive results compared to geometry-based methods without the need for parameter tuning of traditional VO systems. A similar model called UnDeepVO \cite{li2018undeepvo} uses an unsupervised learning technique to estimate the depth of the frames and use it to recover the scale. The depth and scale are obtained by stereo images during training, however, only monocular images are used during the test, making the system monocular.


As we observed in our literature review, there are only a few supervised end-to-end models for 6-DoF pose estimation \cite{wang2020approaches} in the context of visual odometry with imagery data, since most researched approaches are hybrid that still use geometric modeling in their VO components, or they deal with VIO, as can be seen in \cite{qin2022survey, wang2020approaches, neyestani2023survey}. Furthermore, the following surveys do not mention the use of attention mechanisms in the context of end-to-end models \cite{qin2022survey, wang2020approaches,neyestani2023survey, favorskaya2023deep}, that is, they concentrate on works based on CNN and RNN with LSTM to deal with spatio-temporal features. 
Nevertheless, some articles that use the Transformer architecture in visual odometry problems were found outside the surveys. However, these are still hybrid approaches that use the Transformer for multi-modal fusion with LiDAR \cite{sun2023transfusionodom}, relating temporal information from optical flow and CNN modules \cite{ebin2023vit}, and none were found with end-to-end based on space-time self-attention.

Therefore, our work aims to contribute specifically to the area of end-to-end supervised methods in monocular visual odometry. This focus is one of the main reasons we selected DeepVO as a baseline model, considered to be an important and well-established fully supervised end-to-end model for monocular visual odometry.


\subsection{Video understanding}
Video understanding comprehends the recognition and localization of actions or events in videos. A typical task is called action recognition, which consists of assigning a single action to a clip. It is also possible to detect more than one action in videos, by classifying the actions in bounding boxes, similar to object detection problems \cite{feichtenhofer2019slowfast}.

SlowFast networks \cite{feichtenhofer2019slowfast} are high-performance networks for video understanding tasks. It has a fast path that captures motion with high temporal resolution (input with a high frame rate), and a slow path that captures spatial features at a low frame rate. This family of networks might be cumbersome in terms of inference cost, measured by the number of multiply-add operations (FLOPs). This motivated the X3D architectures \cite{feichtenhofer2020x3d} that explore the expansion of a 2D image model along time, space, depth, and width axes. The X3D models achieved competitive performance compared with SlowFast architectures while requiring fewer FLOPs, i.e. being lighter in terms of computational cost. 

Both SlowFast and X3D networks are CNN-based models that do not incorporate attention mechanisms in their architectures. Following the success of the Transformer architecture over convolutional networks in vision tasks, researchers have explored using Transformers to extract spatio-temporal features for video understanding tasks. The TimeSformer \cite{gberta_2021_ICML} model uses the Transformer architecture to extract spatio-temporal features across both space and time, achieving state-of-the-art results on action recognition benchmarks and outperforming previous CNN-based models. It is an extension of the ViT applied to the space-time volume, where the video is converted into a sequence of patches embedded through a linear map with added positional information. Different designs of the self-attention blocks in TimeSformer were investigated: space attention, joint space-time attention, divided space-time attention, sparse local global attention, and axial attention. A competitive architecture is the ViViT \cite{Arnab_2021_ICCV} that explored different ways of extracting spatio-temporal features similar to TimeSformer. 

In our work, we explore the divided space-time attention to estimate the 6-DoF pose in monocular visual odometry tasks, since this architecture is designed for computational efficiency while keeping the accuracy performance compared to the other self-attention blocks. To the best of our knowledge, our work is the first to use a Transformer-based architecture with space-time attention in a supervised end-to-end manner, using a video understanding model to deal with the monocular visual odometry task.

\section{Proposed method} 
\label{sec:method}

This section introduces the proposed method and the pre-processing and post-processing steps of our methodology. The architecture extracts spatio-temporal features with a self-attention mechanism to estimate the 6-DoF camera's pose. As pre-processing, the ground truth's absolute coordinates are converted to relative transforms between time steps. Besides, the ground truth's rotation matrix is represented as Euler angles. The post-processing involves reversing the transformations done in the pre-processing step, as well as adjusting the estimated motions that occur repeatedly due to the overlapped input clips, which is described in detail in Subsection~\ref{subsec:post-proc}.

\subsection{TSformer-VO}

\begin{figure*}[!t]
\centering
\includegraphics[width=0.85\textwidth]{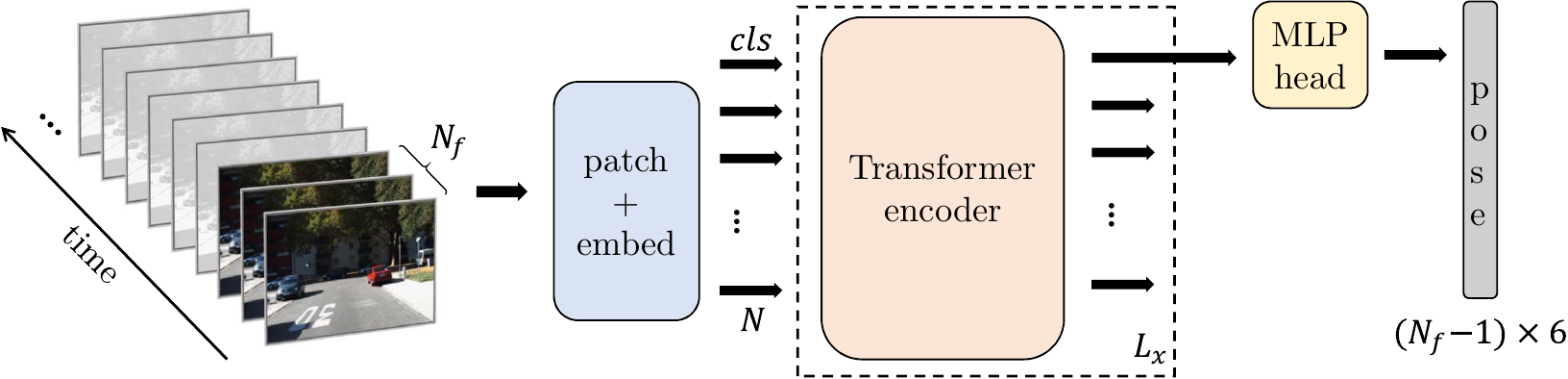}
\caption{TSformer-VO pipeline. The input clips with $N_f$ frames are processed into $N$ patches. Each patch is embedded into tokens and sent to the sequence of Transformer blocks. A special vector called class token (cls) gathers the information from all patches and passes through the final MLP head, outputting the 6-DoF for the $N_f -1$ estimated poses.}
\label{fig:tsformer-vo}
\end{figure*}

The TSformer-VO model is based on the TimeSformer architecture for action recognition \cite{gberta_2021_ICML}. We applied a regression loss to estimate the 6-DoF pose of the input clip and explored the divided space-time self-attention blocks in our context. Fig.~\ref{fig:tsformer-vo} illustrates schematically the steps employed in this architecture.

The input clip $\mathbf{X} \in \mathbb{R}^{N_f \times C \times H \times W}$ consists of $N_f$ frames in the channel first setup, where $C$, $H$, and $W$ are, respectively, the number of channels, the height, and the width of the image frames. Each clip is decomposed into $N$ patches of size $P \times P$, giving $N = HW/P^2$ non-overlapping patches. 
In sequence, each patch is flattened into a vector $\mathbf{x}_{(s, t)} \in \mathbb{R}^{C P^2}$, where $s$ is an index that denotes the patch's spatial location, and $t$ denotes the index over the frames in the clip. Each patch vector is embedded into tokens by a linear learnable map, equivalent to a 2D convolution. The output of this projection is the input token of the Transformer encoder. Those steps of decomposing into patches and embedding into tokens are traditional procedures introduced in \cite{dosovitskiy2020image}. 

Furthermore, we added the learnable positional embedding to each input token, and we follow the BERT Transformer \cite{devlin2018bert} by adding a special learnable vector as the very first input token, known as class token. Although the name classification token is used, in our context we are not dealing with a classification problem, like in image classification and action recognition tasks. However, the idea of keeping this special parameter is to gather the information from all patches, avoiding the bias of choosing the output of a specific patch to pass through the multilayer perceptron (MLP) head. 

Therefore, the final 1D input tokens are denoted as $\mathbf{z}^{l}_{(s, t)} \in \mathbb{R}^{E_d}$, where $E_d$ is the embedding dimension of the flattened patch $\mathbf{x}_{(s, t)}$ at encoding block $l$. The final model has $L_x$ encoder blocks stacked on top of each other. 

There are mainly two different self-attention architectures to extract spatio-temporal features defined in \cite{gberta_2021_ICML}: ``joint space-time'' and ``divided space-time''. In ``joint space-time'' self-attention, all spatio-temporal tokens extracted from the clip are forwarded to the model. This makes the transformer layer relate all pairs of interactions between tokens in space and time together, and therefore the long interactions between tokens throughout the video might require high computational complexity. 

For this reason, due to real-time computation concerns in visual odometry applications, we selected the encoder architecture with the ``divided space-time'' self-attention, which is more efficient in terms of processing complexity compared to the ``joint space-time'' self-attention. As the name suggests, it consists of applying attention along the time axis before the spatial one. Temporal attention uses tokens with the same spatial index, while spatial attention uses tokens from the same frame. 

The MLP head outputs the final estimated poses. Since two frames are required to estimate one pose, $N_f - 1$ poses are estimated from $N_f$ frames. Therefore, the final MLP head has dimension $(N_f -1) \times 6$, considering that we estimate the 6-DoF poses as being a 3D translation and a 3D rotation with Euler angles, described in detail in Subsection~\ref{subsec:pre-proc}. Note that our approach considers the information of all frames in the clip to infer a pose or multiple poses in the clip, depending on the number of frames.

\begin{figure*}[!t]
\centering
\includegraphics[width=0.7\textwidth]{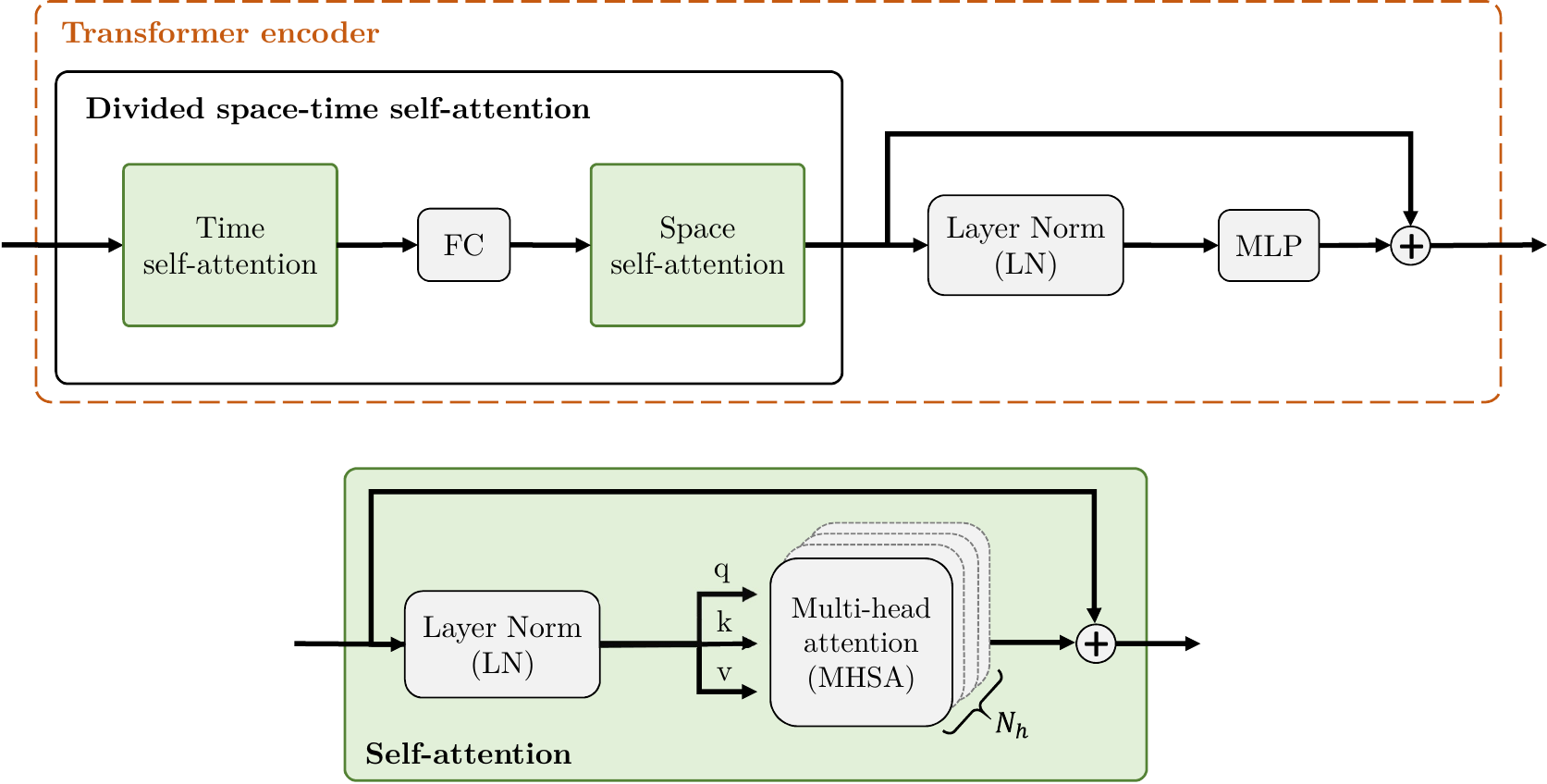}
\caption{Transformer encoder with the divided space-time self-attention architecture. The illustration of the encoder was inspired by \cite{gberta_2021_ICML}.}
\label{fig:transformer-encoder}
\end{figure*}

As shown in Fig.~\ref{fig:transformer-encoder}, both time self-attention and space self-attention of the Transformer encoder have the same architecture depicted in the general self-attention block at the bottom of Fig.~\ref{fig:transformer-encoder}. Furthermore, the encoder blocks comprise the layer normalization (LN), multi-head self-attention (MHSA), residual connections, fully connected layer (FC), and MLP. Their simplified relation for the divided space-time architecture is given as follows:
\begin{equation}
    \begin{split}
        \mathbf{a}^{l}_{t} & =  \text{MHSA}\left(\text{LN}\left(\mathbf{z}_{(s,t)}^{l-1}\right)\right) + \mathbf{z}_{(s,t)}^{l-1}, \\
        \mathbf{a}^{l}_{t_{FC}} & = \text{FC}\left( \mathbf{a}^{l}_{t} \right), \\
        \mathbf{a}^{l}_{s} & = \text{MHSA}\left(\text{LN}\left(\mathbf{a}^{l}_{t_{FC}}\right)\right) + \mathbf{a}^{l}_{t_{FC}}, \\
        \mathbf{z}_{(s,t)}^{l} & = \text{MLP}\left(\text{LN}\left(\mathbf{a}^{l}_{s}\right)\right) + \mathbf{a}^{l}_{s}. \\
    \end{split}
\label{eq:encoder}
\end{equation}

The computation of the query (q), key (k), and value (v), input to the MSHA, is discussed in detail in \cite{vaswani2017attention, gberta_2021_ICML}. The multi-head self-attention has $N_h$ attention heads computed in parallel, each one following Eq.~\eqref{eq:encoder}.

We built the architectures based on the Data Efficient Image Transformer (DEIT) small \cite{deit2021}, that is, with depth $L_x = 12$, MHSA component with $N_h = 6$ heads, and embedding dimension $E_d = 384$. The patch size was $P = 16$. The number of frames $N_f$ in each clip impacts the number of parameters in the model since the output is a dense layer with $6 (N_f - 1)$ neurons. The different models are defined in Table~\ref{tab:model-params} according to their parameters.

\begin{table}[!b]
\caption{TSformer-VO architectures.}
\label{tab:model-params}
\centering
\begin{tabular}{lll}
\hline
\multicolumn{1}{c}{\textbf{Architecture}} & \multicolumn{1}{c}{$N_f$} & \multicolumn{1}{c}{\textbf{\# Parameters}} \\ \hline
\text{TSformer-VO-1} & 2 & 30,657,414 \\
\text{TSformer-VO-2} & 3 & 30,660,108 \\
\text{TSformer-VO-3} & 4 & 30,662,802 \\ \hline
\end{tabular}
\end{table}

\subsection{Pre-processing}
\label{subsec:pre-proc}

Pre-processing is an essential step when the data is labeled with global coordinates. In addition to the standard normalization using the training data statistics, i.e. subtracting the mean and dividing by the standard deviation, the pre-processing also comprises the conversion to relative coordinates and the correct angle notation. 

\subsubsection{Absolute to relative coordinates} 

In visual odometry, the motion $\mathbf{T}_k \in \mathbb{R}^{4\times4}$ is defined as
$$\mathbf{T}_k = \begin{bmatrix}
                \mathbf{R} & \mathbf{t}  \\
                \mathbf{0}   & \mathbf{1}  \\
                \end{bmatrix},  $$
where $k$ is the current time step, $\mathbf{R} \in S \! O(3)$ is the rotation matrix that describes the camera rotation, and $\mathbf{t} \in \mathbb{R}^{3 \times 1}$ is the translation vector. Both the rotation matrix and the translation vector depict the motion from time step $k-1$ to $k$.

The world absolute coordinates might not be useful for pose estimation, since we want to estimate the relative motion between consecutive frames. Let us assume we have the ground truth 3D local coordinates motion $\mathbf{T}_{k-1}$ and $\mathbf{T}_{k}$. The pose of the camera at instant $k$ relative to instant $k-1$, denoted as $\mathbf{T}_{k,k-1}$, is given by:
\begin{equation}
    \mathbf{T}_{k, k-1} = \mathbf{T}_{k-1}^{-1} \mathbf{T}_{k}.
\label{eq:abs2rel}    
\end{equation}

\subsubsection{Rotation matrix to Euler angles} 

The rotations about the $x$, $y$, and $z$ axes are referred to as roll, pitch, and yaw, respectively, and are denoted as $\phi$, $\theta$, and $\psi$, respectively. The rotation matrix $\mathbf{R}$ can be expressed in terms of these rotations as follows:

\begin{multline}
\mathbf{R}(\phi, \theta, \psi) = 
            \begin{bmatrix}
            r_{11}  &  r_{12} & r_{13} \\
            r_{21}  &  r_{22} & r_{23} \\
            r_{31}  &  r_{32} & r_{33} \\
            \end{bmatrix} \\
            = \begin{bmatrix}
            c_{\psi}c_{\theta}  & \; c_{\psi}s_{\theta}s_{\phi}-s_{\psi}c_{\phi} & \; s_{\psi} s_{\phi} + c_{\psi} s_{\theta} c_{\phi} \\
            s_{\psi} c_{\theta} & \; c_{\psi}c_{\phi}+s_{\psi}s_{\theta}s_{\phi} & \; s_{\psi}s_{\theta}c_{\phi} - c_{\psi}s_{\phi} \\
            -s_{\theta}  & \; c_{\theta}s_{\phi}  & \;  c_{\theta}c_{\phi} \\
            \end{bmatrix},
\end{multline}
where $c_{\alpha} = \cos{(\alpha)}$ and  $s_{\alpha} = \sin{(\alpha)}$, $\alpha \in \{\phi, \theta, \psi\}$. This representation follows the ZYX Euler parameterization, which yields the following Euler angles \cite{lynch2017modern}:
\begin{equation}
    \begin{split}
        \phi & = \atantwo(r_{21}, r_{11}), \\
        \theta & = \atantwo\left(-r_{31}, \sqrt{r_{11}^2 + r_{21}^2}\right), \\
        \psi & = \atantwo(r_{32}, r_{33}), \\
    \end{split}
\end{equation}
where $\atantwo(y,x)$ is the two-argument $\arctan$ that returns the angle between the vector $(x, y)$ and the $x$ axis. 

\subsection{Post-processing}
\label{subsec:post-proc}
Post-processing involves recovering the global coordinates given the relative coordinates predicted by the model. For this purpose, the reverse operation of Eq.~\eqref{eq:abs2rel} can be applied, as well as transforming the Euler angles back to the rotation matrix. Moreover, the training statistics used in normalizing the data should be reemployed to retrieve the denormalized estimated poses.

Furthermore, given the way we designed our problem, poses at the same time step can appear in different input clips. In our approach, we use input clips with $N_f$ frames overlapped with $N_f-1$ frames, as if a sliding window spans the video with stride 1. In other words, the clips act as a circular buffer of size $N_f$, and at each time step the oldest frame goes away and a new frame is appended to the clip. 

\begin{figure}[!t]
\centering
\includegraphics[width=0.48\textwidth]{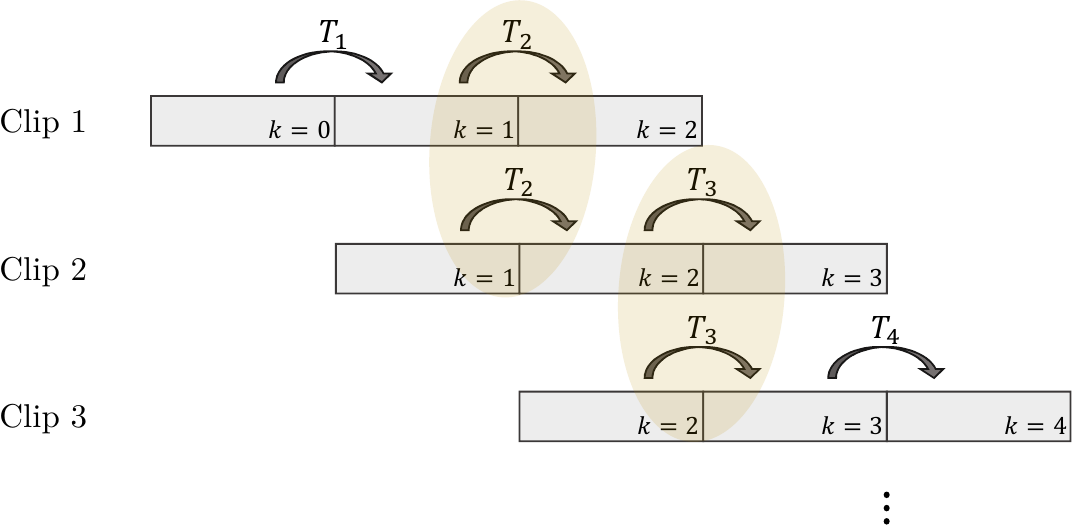}
\caption{Visualization of the repeated motions, highlighted in yellow, for the particular case of $N_f = 3$ with $2$ overlapped frames.}
\label{fig:repeated_motions}
\end{figure}

This overlap leads to the same pose being estimated with different input clips. Hence, we perform a minor adjustment by averaging all $\mathbf{T}_k$ motions over all clips where $\mathbf{T}_k$ appears given a time instant $k$. Fig.~\ref{fig:repeated_motions} helps with the visualization of the repeated motions in consecutive clips. Note that Fig.~\ref{fig:repeated_motions} shows a particular case for $N_f = 3$ with 2 overlapped frames. The repeated motions are highlighted in light yellow.

Also, calculating the motion this way introduces a delay where the system needs to wait for other clips to appear in order to output a predicted motion for a time step. This trade-off between the introduced delay and averaging the redundancy over the repeated motions does not introduce significant delays considering the small number of frames in the clips.

\section{Experiments}
\label{sec:experiment}

The data and training setups used in our experiments are described in this section. 

\subsection{Experimental setup}

In the following, we describe in detail the dataset, and the metrics we computed to evaluate our approach against the state-of-the-art. We also detail the training hyperparameters and the computational setup.

\subsubsection{Dataset}

We conduct our experiments with the KITTI odometry dataset \cite{Geiger2012CVPR}, which is a benchmark for the development and evaluation of VO algorithms. The data is recorded at 10 frames per second by a stereo system mounted on a moving vehicle, capturing images while riding through streets and roads. Therefore, the scenes might contain pedestrians, parked and moving cars, bicycles, etc. In our case, since we deal with monocular VO, we only consider the images acquired by the left camera. The entire data has 22 sequences: 11 with ground truth poses provided by a GPS for developing and evaluating the method, and 11 with no ground truth available. The length of the sequences is not equal, and the car speed varies from 0 to 90 km/h, making the VO challenging in high-speed and steep curve situations.

\subsubsection{Evaluation metrics}

We evaluate the methods using the KITTI odometry metrics, as defined below:

\begin{itemize}
    \item ATE: Absolute Trajectory Error, measured in meters. It represents the root mean squared error (RMSE) between the estimated camera poses and the ground truth.
    \item RPE: Relative Pose Error for both rotation and translation. This is calculated frame-to-frame, where rotation is measured in degrees ($^{\circ}$), and translation is measured in meters (m).
    \item $t_{err}$: Average translational error, measured as a percentage (\%).
    \item $r_{err}$: Average rotational error, measured in degrees per 100 meters ($^{\circ}$/100 m).
\end{itemize}

The ATE evaluates the root mean squared error between the ground truth and the predicted translational component of the pose \cite{sturm2012benchmark}. Let $\mathbf{t}_k$ be the ground truth translation vector of the frame at instant $k$, and $\hat{\mathbf{t}}_k$ the predicted translation vector. The ATE is defined as:
\begin{equation}
\text{ATE} = \sqrt{\frac{1}{L} \sum_{k=1}^L \|\mathbf{t}_k - \hat{\mathbf{t}}_k\|^2_2}
\end{equation}
where $\left\| \cdot \right\|_2$ denotes the Euclidean norm, and $L$ is the total number of frames in the sequence.

The RPE measures the error between the relative motion of two consecutive frames, comparing the ground truth and predicted relative poses. As proposed by Geiger \textit{et al.} \cite{Geiger2012CVPR}, the RPE is computed separately for rotation ($\text{RPE}_{\text{rot}}$) and translation ($\text{RPE}_{\text{trans}}$). The relative pose error at frame $k$ is defined as:
\begin{equation}
    \mathbf{\epsilon}_k = \mathbf{T}_{k,k-1}^{-1} \hat{\mathbf{T}}_{k,k-1},
\end{equation}
where $\mathbf{T}_{k,k-1}$ is the ground truth relative pose, as defined in Eq.~\eqref{eq:abs2rel}, and $\hat{\mathbf{T}}_{k,k-1}$ is its predicted counterpart.

For the translational component ($\text{RPE}_{\text{trans}}$), we calculate the average relative pose error over all frames, as follows:
\begin{equation}
    \text{RPE}_{\text{trans}} = \frac{1}{L-1} \sum_{k=1}^{L-1} \|\mathbf{\epsilon}_{t,k}\|_2,
\end{equation}
where $\mathbf{\epsilon}_{t,k}$ represents the translational component of the relative pose error $\mathbf{\epsilon}_k$.

Similarly, for the rotational component, the average relative pose error is given by:
\begin{equation}
    \text{RPE}_{\text{rot}} = \frac{1}{L-1} \sum_{k=1}^{L-1} \arccos \left( \frac{\mathrm{Tr}(\mathbf{\epsilon}_{R, k}) - 1}{2} \right),
\end{equation}
where $\mathbf{\epsilon}_{R, k}$ represents the rotational component of the relative pose error, and $\mathrm{Tr}(\cdot)$ is the trace of the rotation matrix of the relative pose error.

The metrics $t_{err}$ and $r_{err}$ are computed by averaging the translational and rotational errors over subsequences of varying lengths. Following \cite{Geiger2012CVPR}, the errors are calculated for subsequences of length $(100, 200, \dots, 800)$ meters. 

Monocular methods typically suffer from scale ambiguity when reconstructing real-world scale. Prior works apply a transformation optimization to align the predicted poses with the ground truth. Accordingly, we applied a 7-DoF optimization during validation, as is common in the literature \cite{zhan2020visual, orbslam2}. The final metrics, after optimization, were computed using the Python KITTI evaluation toolbox\footnote{https://github.com/Huangying-Zhan/kitti-odom-eval}.

\subsubsection{Loss function}

The loss function used in this work is the mean squared error (MSE) between each predicted element and its ground truth target, defined as follows:

\begin{equation}
    \mathcal{L}_{\textsc{}{MSE}} = \frac{1}{6 B_s (N_f -1)} \sum_{n=1}^{B_s}\sum_{i=1}^{6(N_f -1)}{\left(\mathbf{y}_{i,n} - \hat{\mathbf{y}}_{i,n}\right)^2}, \label{eq:loss}
\end{equation}
where $\mathbf{y}_{i,n}$ is the flattened 6-DoF ground truth's relative poses of batch $n$, $i$ is the $i$-th element of the flattened column vector, and $\hat{\mathbf{y}}_{i,n}$ is its prediction by the model. For batch processing, the final loss is reduced by the mean over all $B_s$ batch elements, as seen in Eq.~\eqref{eq:loss}. 

\subsubsection{Training strategy}
\label{subsec:training}
Out of the 11 KITTI sequences with ground truth, we used sequences 00, 02, 08, and 09 as training data, and 01, 03, 04, 05, 06, 07, and 10 as test data. We follow the choice made in \cite{wang2017deepvo} to conduct a fair comparison. Although only four sequences are used for training, they contain the largest recorded trajectories in the dataset. All the frames were resized to $192\times640$ to make the height and width dimensions multiples of the patch size ($P=16$) while keeping the original aspect ratio of the dataset.

The input clips with $N_f$ frames, $N_f \in \{2, 3, 4\}$, are sampled from the KITTI dataset. We sampled the clips using a sliding window of size $N_f$ and stride 1. This means that consecutive clips have $N_f - 1$ overlapped frames. Next, we shuffle the sampled clips and organize them in batches of size 4. By doing this, we try to ensure that clips in the same batch are not sequential to each other. The batch size was picked according to our GPU memory capacity.

As validation data, we randomly select 10\% of the training clips, so that the validation set has the same distribution as the training data. We compute the validation loss to do hyperparameter tuning and save the best model, i.e. the model with minimum validation loss during training.

\begin{figure}[!t]
\centering
\includegraphics[width=0.47\textwidth]{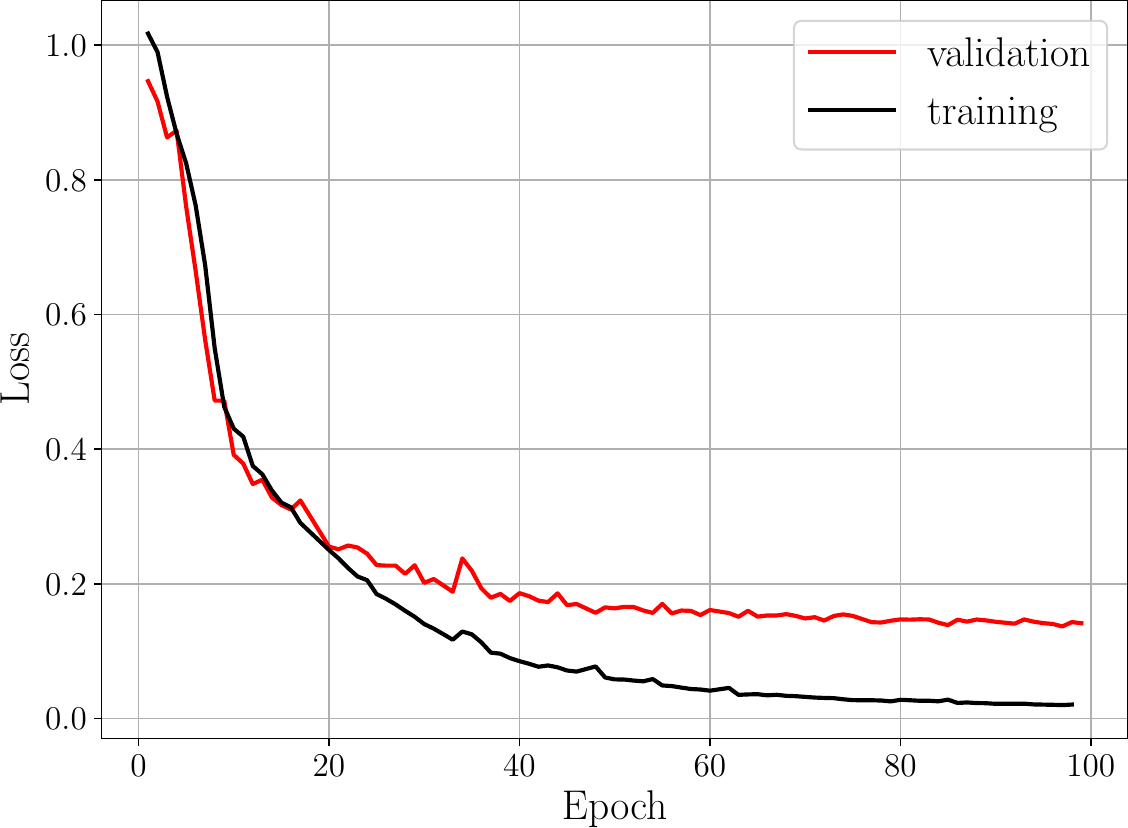}
\caption{Training and validation loss curves of TSformer-VO-1 architecture.}
\label{fig:loss}
\end{figure}

The training procedure and the architectures were implemented with PyTorch 1.10. Computations were performed with a computer with an Intel i9-7900X CPU 3.3GHz CPU and a GeForce GTX 1080 Ti GPU with 11GB VRAM. We run experiments from scratch for 100 epochs with the Adam optimization to minimize the loss function. We set the learning rate to $1\times 10^{-5}$ and use the default values for the other parameters in the Adam algorithm. Fig.~\ref{fig:loss} shows the training and validation loss convergence curves obtained by training the architecture TSformer-VO-1.

\subsection{Results}

The qualitative and quantitative results of our TSformer-VO models are presented in the following, evidencing the potential of our proposed method. We also show the visualization of the learned spatio-temporal attention on the KITTI images, and the total computational time of our approach.

\subsubsection{Comparison with the state-of-the-art}
We conduct experiments to evaluate our approach by comparing it to other classical models in the literature. We evaluate the performance of our method by comparing it to the visual odometry component of ORB-SLAM3, which is a competitive popular geometry-based method, and DeepVO, which is a famous end-to-end deep learning-based method. 

In addition, we used the official C++ implementation of ORB-SLAM3\footnote{https://github.com/UZ-SLAMLab/ORB\_SLAM3} and the unofficial PyTorch implementation of DeepVO\footnote{https://github.com/ChiWeiHsiao/DeepVO-pytorch}. It is worth mentioning that the authors of DeepVO did not release the official code implementation. Therefore, we were cautious to choose the highest-rated unofficial PyTorch version available online on GitHub, also used in other related works \cite{bruno2021lift, siyuan2022, kaygusuz2021mdn, kaygusuz2022aft}. Furthermore, we do not apply loop closure on ORB-SLAM3 to keep a fair comparison between the algorithms, since DeepVO and TSformer-VO do not use loop detection to perform a pose-graph optimization. 

The architectures defined in Table~\ref{tab:model-params} were then trained and evaluated. The quantitative results are given in Table~\ref{tab:metrics}. The best values of each metric for each sequence are highlighted in bold, and the second-best values are underlined.

\begin{table*}
\centering
\caption{Quantitative results of the selected models for the 11 KITTI sequences with ground truth. The best values are highlighted in bold, and the second-best are underlined.}
\label{tab:metrics}
\begin{tabular}{cllllllll}
\hline
\multirow{2}{*}{}                                                             & \multicolumn{1}{c}{\multirow{2}{*}{\textbf{Method}}} & \multicolumn{7}{c}{\textbf{Sequence}}                                                                                                                                                                                                       \\ \cline{3-9} 
                                                                              & \multicolumn{1}{c}{}                                 & \multicolumn{1}{c}{\textbf{01}} & \multicolumn{1}{c}{\textbf{03}} & \multicolumn{1}{c}{\textbf{04}} & \multicolumn{1}{c}{\textbf{05}} & \multicolumn{1}{c}{\textbf{06}} & \multicolumn{1}{c}{\textbf{07}} & \multicolumn{1}{c}{\textbf{10}} \\ \hline
\multirow{5}{*}{\begin{tabular}[c]{@{}c@{}}$t_{err}$\\ (\%)\end{tabular}}     & ORB-SLAM3                                            & 112.198                         & \textbf{1.269}                  & \textbf{1.389}                  & 59.677                          & \textbf{17.592}                 & \textbf{12.727}                 & \textbf{5.672}                  \\
                                                                              & DeepVO                                               & 99.047                          & 85.203                          & 18.812                          & 48.869                          & 52.220                          & 61.080                          & 114.054                         \\
                                                                              & TSformer-VO-1                                        & 37.322                          & 14.731                          & 8.242                           & \textbf{9.623}                  & 25.053                          & \uline{17.013}                    & 15.459                          \\
                                                                              & TSformer-VO-2                                        & \textbf{33.400}                 & 14.444                          & 6.854                           & \uline{10.735}                    & \uline{17.703}                    & 23.205                          & \uline{13.713}                    \\
                                                                              & TSformer-VO-3                                        & \uline{35.504}                    & \uline{12.858}                    & \uline{5.664}                     & 12.588                          & 28.973                          & 22.959                          & 16.072                          \\ \hline
\multirow{5}{*}{\begin{tabular}[c]{@{}c@{}}$r_{err}$\\ (º/100m)\end{tabular}} & ORB-SLAM3                                            & \textbf{1.370}                  & \textbf{0.388}                  & \textbf{0.263}                  & 30.851                          & \textbf{0.430}                  & \textbf{1.889}                  & \textbf{1.508}                  \\
                                                                              & DeepVO                                               & 12.930                          & 24.602                          & 7.176                           & 35.305                          & 33.569                          & 59.860                          & 26.632                          \\
                                                                              & TSformer-VO-1                                        & 5.321                           & 6.988                           & 4.849                           & \textbf{3.629}                  & 8.443                           & \uline{6.361}                     & \uline{4.670}                     \\
                                                                              & TSformer-VO-2                                        & 6.251                           & 6.129                           & 3.556                           & \uline{4.002}                     & \uline{5.622}                     & 9.992                           & 5.111                           \\
                                                                              & TSformer-VO-3                                        & \uline{5.192}                     & \uline{5.756}                     & \uline{3.492}                     & 5.133                           & 8.838                           & 11.544                          & 5.161                           \\ \hline
\multirow{5}{*}{\begin{tabular}[c]{@{}c@{}}ATE\\ (m)\end{tabular}}            & ORB-SLAM3                                            & 524.966                         & \textbf{0.883}                  & \textbf{1.353}                  & 84.239                          & \textbf{49.744}                 & \uline{18.079}                    & \textbf{9.370}                  \\
                                                                              & DeepVO                                               & \textbf{68.258}                 & 21.021                          & 5.648                           & 54.860                          & 88.468                          & \textbf{7.961}                  & 22.755                          \\
                                                                              & TSformer-VO-1                                        & \uline{126.225}                   & 16.623                          & 4.750                           & \textbf{46.890}                 & 78.820                          & 32.883                          & 22.975                          \\
                                                                              & TSformer-VO-2                                        & 209.038                         & 14.737                          & 4.244                           & \uline{54.688}                    & \uline{50.519}                    & 36.059                          & \uline{21.131}                    \\
                                                                              & TSformer-VO-3                                        & 160.546                         & \uline{14.152}                    & \uline{3.057}                     & 61.387                          & 88.314                          & 31.490                          & 22.696                          \\ \hline
\multirow{5}{*}{\begin{tabular}[c]{@{}c@{}} $\text{RPE}_{\text{trans}}$ \\ (m)\end{tabular}}            & ORB-SLAM3                                            & 3.036                           & \textbf{0.025}                  & \textbf{0.054}                  & 0.845                           & \uline{0.280}                     & \textbf{0.124}                  & \textbf{0.060}                  \\
                                                                              & DeepVO                                               & 2.277                           & 0.666                           & 0.337                           & 0.605                           & 0.838                           & 0.524                           & 1.002                           \\
                                                                              & TSformer-VO-1                                        & 1.007                           & 0.134                           & 0.086                           & \uline{0.142}                     & 0.326                           & \uline{0.136}                     & 0.148                           \\
                                                                              & TSformer-VO-2                                        & \textbf{0.751}                  & \uline{0.128}                     & \uline{0.083}                     & 0.144                           & \textbf{0.209}                  & 0.143                           & \uline{0.145}                     \\
                                                                              & TSformer-VO-3                                        & \uline{0.953}                     & 0.134                           & 0.112                           & \textbf{0.139}                  & 0.404                           & 0.162                           & 0.154                           \\ \hline
\multirow{5}{*}{\begin{tabular}[c]{@{}c@{}}$\text{RPE}_{\text{rot}}$\\ (º)\end{tabular}}            & ORB-SLAM3                                            & \textbf{0.157}                  & \uline{0.169}                     & \uline{0.111}                     & \textbf{0.111}                  & \textbf{0.126}                  & 0.358                           & 0.397                           \\
                                                                              & DeepVO                                               & 0.624                           & 0.567                           & 0.171                           & 0.716                           & 0.610                           & 0.940                           & 0.746                           \\
                                                                              & TSformer-VO-1                                        & \uline{0.284}                     & 0.230                           & 0.155                           & 0.213                           & \uline{0.208}                     & \uline{0.230}                     & \uline{0.285}                     \\
                                                                              & TSformer-VO-2                                        & 0.751                           & \textbf{0.128}                  & \textbf{0.083}                  & \uline{0.144}                     & 0.209                           & \textbf{0.143}                  & \textbf{0.145}                  \\
                                                                              & TSformer-VO-3                                        & 0.294                           & 0.292                           & 0.173                           & 0.279                           & 0.264                           & 0.322                           & 0.333                           \\ \hline
\end{tabular}
\end{table*}

To conduct a fair comparison among the methods, we ignored the training sequences 00, 02, 08, and 09 as the learning nature of DeepVO and TSformer-VO is supervised, while ORB-SLAM3 is strictly geometric without a prior training step. That stated, we can see that the TSformer-VO model is competitive with the other state-of-the-art methods analyzed. ORB-SLAM3 is still superior considering the rotation metrics, such as $r_{err}$ and $\text{RPE}_{\text{rot}}$. However, TSformer-VO showed competitive results in translations, especially evident in sequences 01, 05, and 06 regarding the metrics $t_{err}$ and $\text{RPE}_{\text{trans}}$. 

Comparing algorithms of the same nature, i.e. based on deep learning, the three trained TSformer-VO architectures were clearly superior to the DeepVO method. This shows that transformer-based architectures can also provide high performance in visual odometry tasks. As an example, the DeepVO only shows scores higher than TSformer-VO-2 for the ATE of sequences 01, 07, and 09, and $\text{RPE}_{\text{rot}}$ for sequence 01. For all the remaining metrics and sequences, the TSformer-VO-2 outperforms the DeepVO.

\begin{figure*}[!t]
\centering
\includegraphics[width=0.78\textwidth]{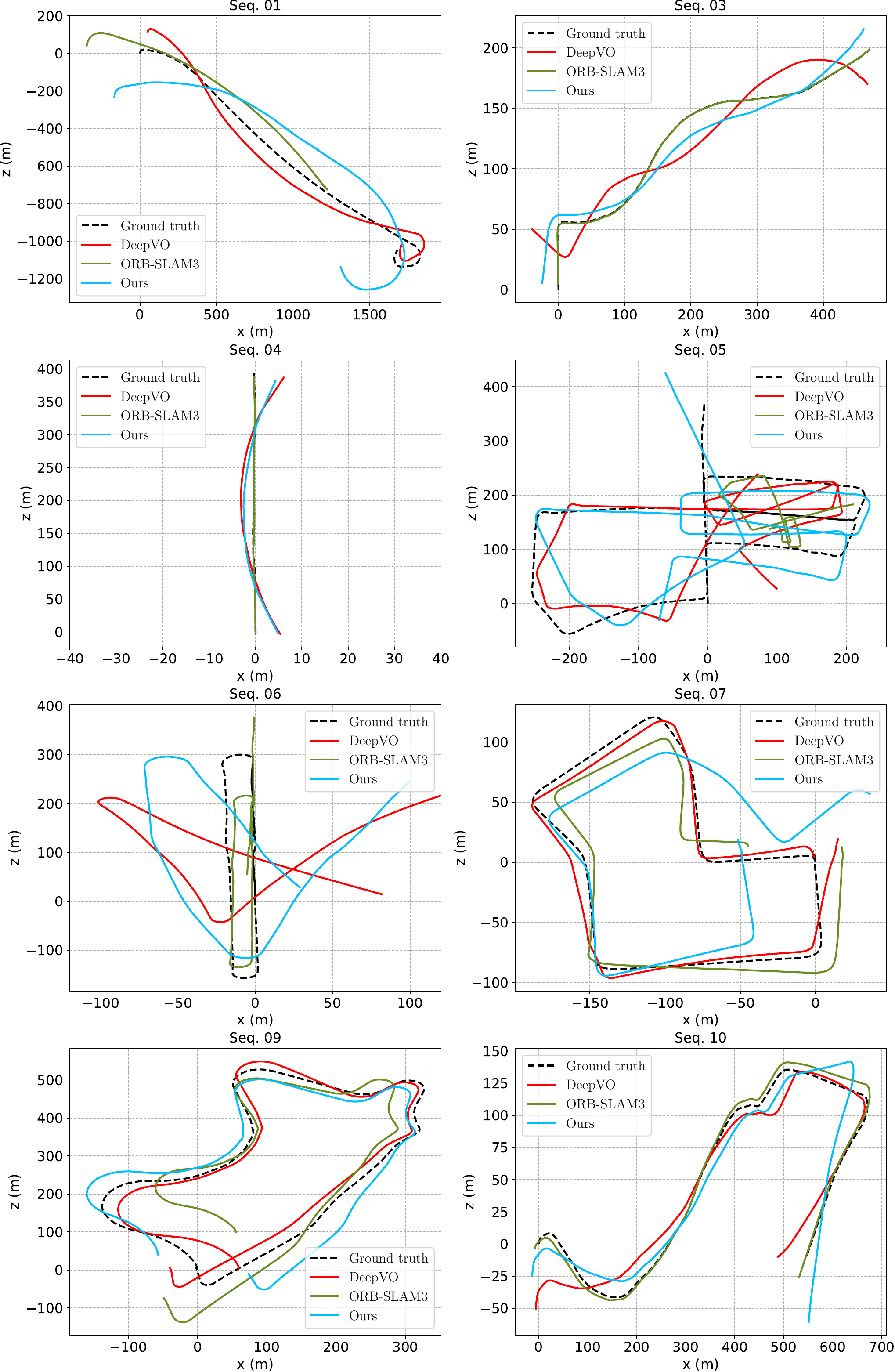}
\caption{Trajectories obtained by the ORB-SLAM3(\colorindicator{tab:olivedrab}), DeepVO (\colorindicator{tab:r}), and TSformer-VO-2 (\colorindicator{tab:deepskyblue}), compared with the ground truth (\colorindicator{tab:k}) in sequences 01, 03, 04, 05, 06, 07, 09, and 10 of the KITTI odometry dataset. The depicted sequences belong to the test set, except sequence 09, and the trajectories are obtained under the 7-DoF alignment.}
\label{fig:traj_results}
\end{figure*}

A qualitative analysis is shown in Fig.~\ref{fig:traj_results}. The predicted trajectories are displayed on top of each other together with the expected ground truth.

It is worth noting that ORB-SLAM3 had the worst translational performance in sequence 01 mainly due to the high-speed scenario in this sequence, hindering the tracking of features along the frames with classical matching and tracking algorithms. This issue is not found in DeepVO and TSformer-VO, revealing an advantage of deep learning-based methods over geometry-based ones in high-speed scenarios.

\begin{figure*}[!t]
\centering
\includegraphics[width=0.97\textwidth]{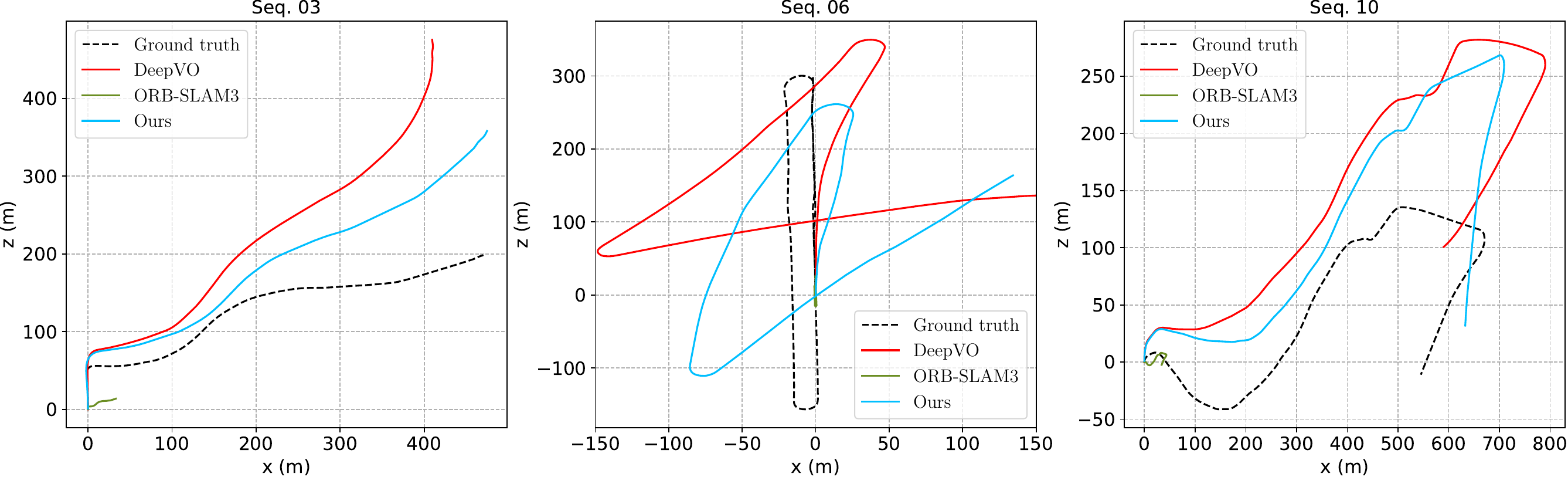}
\caption{Trajectories without the 7-DoF alignment obtained by the ORB-SLAM3(\colorindicator{tab:olivedrab}), DeepVO (\colorindicator{tab:r}), and TSformer-VO-2 (\colorindicator{tab:deepskyblue}), compared with the ground truth (\colorindicator{tab:k}) in sequences 03, 06, and 10 of the KITTI odometry dataset.}
\label{fig:traj_results_without}
\end{figure*}

Furthermore, we observed in our experiments that the DeepVO and TSformer-VO methods are able to estimate better poses in terms of translation without the 7-DoF alignment required by the ORB-SLAM3, as shown in Fig.~\ref{fig:traj_results_without}. This result is favorable for the choice of deep learning-based methods over geometry-based methods, as deep learning indirectly learns the scale in the data during training. However, learning the scale information from the training data may affect their generalization to new datasets, and a scale adjustment may be required to achieve accurate estimation.

\subsubsection{Visualization of the learned space-time attention}

\begin{figure*}[!t]
\centering
\includegraphics[width=0.99\textwidth]{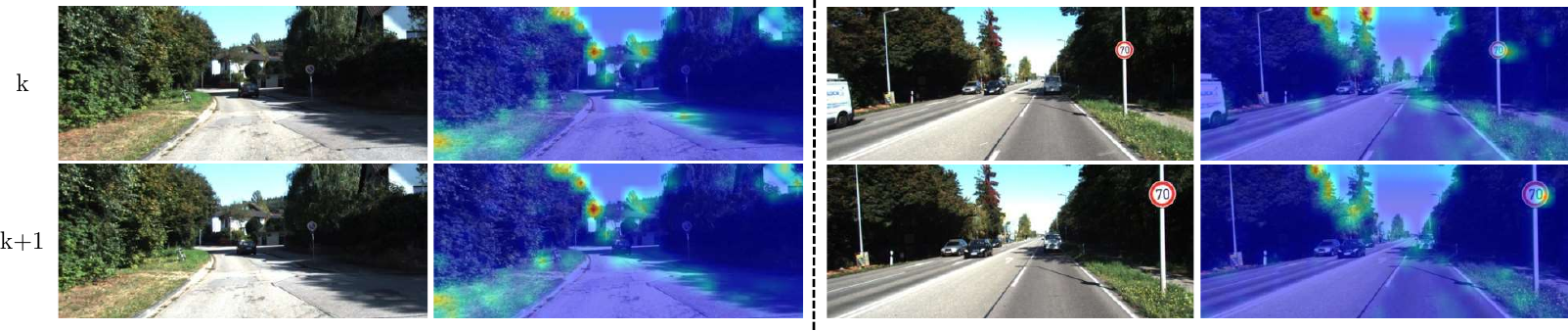}
\caption{Visualization of the learned space-time attention in the context of visual odometry using the KITTI dataset. The more red, the higher the computed attention, and the more blue, the lower the attention.}
\label{fig:attn_map}
\end{figure*}

Understanding what the network is learning might be valuable information to further improve the method. The learned space-time attention can be visualized with the Attention Rollout introduced in \cite{abnar2020quantifying}. This method shows which part of the input is considered important to the network after training. The idea is to propagate the attention weights through the layers quantifying the flow of information. Fig.~\ref{fig:attn_map} shows the visualization of the space-time attention for examples of the KITTI dataset.

Fig.~\ref{fig:attn_map} illustrates an example in which the learned attention focuses on the static scene, ignoring moving objects, such as cars. The learned attention disregards both cars moving in the same direction as the camera and those moving in the opposite direction. This behavior was frequently observed in the KITTI sequences, indicating the network's ability to ignore moving objects and extract relevant features from static scenarios. In addition, differently from classical approaches based on keypoint detectors, the learned space-time attention has blob shapes, giving preference to regions instead of corners. This can be advantageous, as it is easier to track larger objects rather than single keypoints given the number of pixels available in the objects.

\subsubsection{Computational time for TSformer-VO}
Considering the importance of real-time processing for visual odometry, we measured the inference time of our proposed models. We computed the mean computational time and its standard deviation required for our models to estimate the poses in 1100 clips. Table~\ref{tab:runtimes} shows the average inference time for each model, using the computer described in Subsection~\ref{subsec:training}.

\begin{table}
\centering
\caption{Comparison of inference times for TSformer-VO models per clip.}
\label{tab:runtimes}
\begin{tblr}{
  column{2} = {c},
  column{3} = {c},
  hline{1-2,5} = {-}{},
}
\textbf{Model}  & {\textbf{mean }\\\textbf{ (ms)}} & {\textbf{std. dev.}\\\textbf{ (ms)}} \\
TSformer-VO-1   & 20.346                           & 0.032                                \\
TSformer-VO-2   & 28.840                           & 0.505                                \\
TSformer-VO-3   & 37.877                           & 0.822                                
\end{tblr}
\end{table}

The pre- and post-processing require on average $3.369$~ms and $0.031$~ms, respectively, to process a pair of frames. Note that for the test case, the pre-processing step consists only of data normalization and resizing. For a system operating at 10 frames per second, as in the KITTI benchmark, the sample time to capture one frame is $100$ ms. Table~\ref{tab:runtimes} shows that our approach requires an inference time between $20$~ms and $40$~ms on average with our computational setup. Therefore, regarding processing time, even our largest model (TSformer-VO-3) is capable of real-time application once the total processing time is lower than the sample time. 

Notice that despite our method requiring $N_f$ frames, we do not need to wait for $N_f$ frames to be captured before one inference, since we can keep a buffer with the last $N_f-1$ frames and complete it with the current frame. Then, as a new frame arrives, the oldest one is dropped from this buffer while the new frame is added. The average motion computed between the relative pose estimates suggested in Subsection~\ref{subsec:post-proc} as a post-processing step is advised for high-accuracy applications. However, for low-latency applications, such as control systems, this average process may be skipped to avoid introducing delay.

\subsubsection{Ablation study}

During the model development phase, we performed an ablation study to determine the optimal configuration of the ViT model. We began by testing architectures based on standard ViT designs, systematically varying the number of heads in the MHSA, embedding dimension, and Transformer depth. The specific architectures used in our experiments are detailed in Table~\ref{tab:deit_sizes}.

\begin{table}
\centering
\caption{ViT architectures employed during the development phase.}
\label{tab:deit_sizes}
\begin{tabular}{ccccc}
\hline
\textbf{ViT model} & \multicolumn{1}{l}{$P$} & \multicolumn{1}{l}{$E_d$} & \multicolumn{1}{l}{$L_x$} & \multicolumn{1}{l}{$N_h$} \\ \hline
tiny & 16 & 192 & 12 & 3 \\
small & 16 & 384 & 12 & 6 \\
base & 16 & 768 & 12 & 12 \\ \hline
\end{tabular}
\end{table}
Using these configurations, we conducted a comparative analysis to identify the most suitable model, taking into account both performance and hardware constraints. Table~\ref{tab:deit_ablation} presents the average absolute trajectory error and relative pose error across all the test sequences, providing a single metric for evaluation over the test sequences.

\begin{table}
\centering
\caption{Average (avg) ATE and $\text{RPE}_{\text{trans}}$ for the tiny, small and base ViT architectures.}
\label{tab:deit_ablation}
\begin{tabular}{lllll}
\hline
\textbf{Model}     & $E_d$ & $N_h$ & avg ATE (m) & avg $\text{RPE}_{\text{trans}}$ (m) \\ \hline
ViT-tiny & 192   & 3     & 171.48      & 0.831       \\
ViT-small & 384   & 6     & 140.418     & 0.634       \\
ViT-base  & 768   & 12    & 145.426     & 0.518       \\ \hline
\end{tabular}
\end{table}
As shown in Table~\ref{tab:deit_ablation}, the tiny model exhibited the poorest performance. The difference between the small and base models was minimal, indicating that the small model strikes a balance between computational efficiency and performance. This choice is particularly relevant for real-time applications, where inference speed is crucial.

To further refine our approach, we explored the use of pretrained models. Leveraging the architectures in Table~\ref{tab:deit_sizes}, we evaluated the impact of initializing the network with a model pretrained on ImageNet versus training from scratch. Due to faster training times, we conducted these experiments using the tiny model, while keeping unchanged all the other hyperparameters. Table~\ref{tab:deit_pretrained} summarizes the results.

\begin{table}
\centering
\caption{Average (avg) ATE and $\text{RPE}_{\text{trans}}$ for the tiny ViT architecture with and without pretrained ViT.}
\label{tab:deit_pretrained}
\begin{tabular}{llll}
\hline
\textbf{Model} & \textbf{Pretrained} & avg ATE (m) & avg $\text{RPE}_{\text{trans}}$ (m) \\ \hline
ViT-tiny      & False               & 171.48      & 0.831       \\
ViT-tiny       & True                & 186.251     & 3.609       \\ \hline
\end{tabular}
\end{table}

The results in Table~\ref{tab:deit_pretrained} indicate that training the model from scratch produced better outcomes in our specific task. Pretrained models on ImageNet, despite their advantages in generalization and low-level feature extraction capabilities, did not translate effectively to our domain, likely due to the significant domain shift between the datasets.

\section{Conclusion}\label{sec:conclusion}
In this work, we presented an end-to-end supervised model based on Transformer for monocular visual odometry tasks. The features are extracted by a spatio-temporal attention mechanism, and all the 6-DoF camera's poses of the input video clip are estimated by a MLP regressor.

Finally, our experiments have shown that a Transformer-based model originally created for video understanding problems can predict the 6-DoF camera's pose given sequential images as input. We trained and evaluated different architectures varying the last dense layer and the number of frames of the input clips. 

Our results showed that the trained TSformer-VO models achieved competitive results in the KITTI dataset when compared to well-established methods based on deep learning and geometry, respectively DeepVO and ORB-SLAM3. Considering only the deep learning-based method, the rotation and translation metrics support that our approach was superior to the DeepVO implementation widely accepted in the community. Furthermore, we observed that the learned spatio-temporal attention is mostly drawn to the static scene, which is desirable in visual odometry tasks. We also show that our approach enables real-time processing even for our largest model. 

In addition, despite the relatively small size of the KITTI dataset, our model successfully estimated camera poses with performance on par with state-of-the-art methods. Given the nature of deep learning, larger datasets are expected to further enhance the model’s learning capabilities and accuracy, potentially through transfer learning techniques and additional data sources. Future research could explore the robustness of the TSformer-VO model under more diverse environmental conditions, such as varying lighting, dynamic objects, and indoor settings. Investigating unsupervised learning approaches, similar to \cite{li2018undeepvo}, along with experimenting with novel cost functions that incorporate motion consistency in overlapping video clips, may also be promising directions. These extensions could significantly improve the model’s generalization and applicability to more complex and realistic visual odometry tasks. 


\begin{IEEEbiography}[{\includegraphics[width=1in,height=1.25in,clip,keepaspectratio]{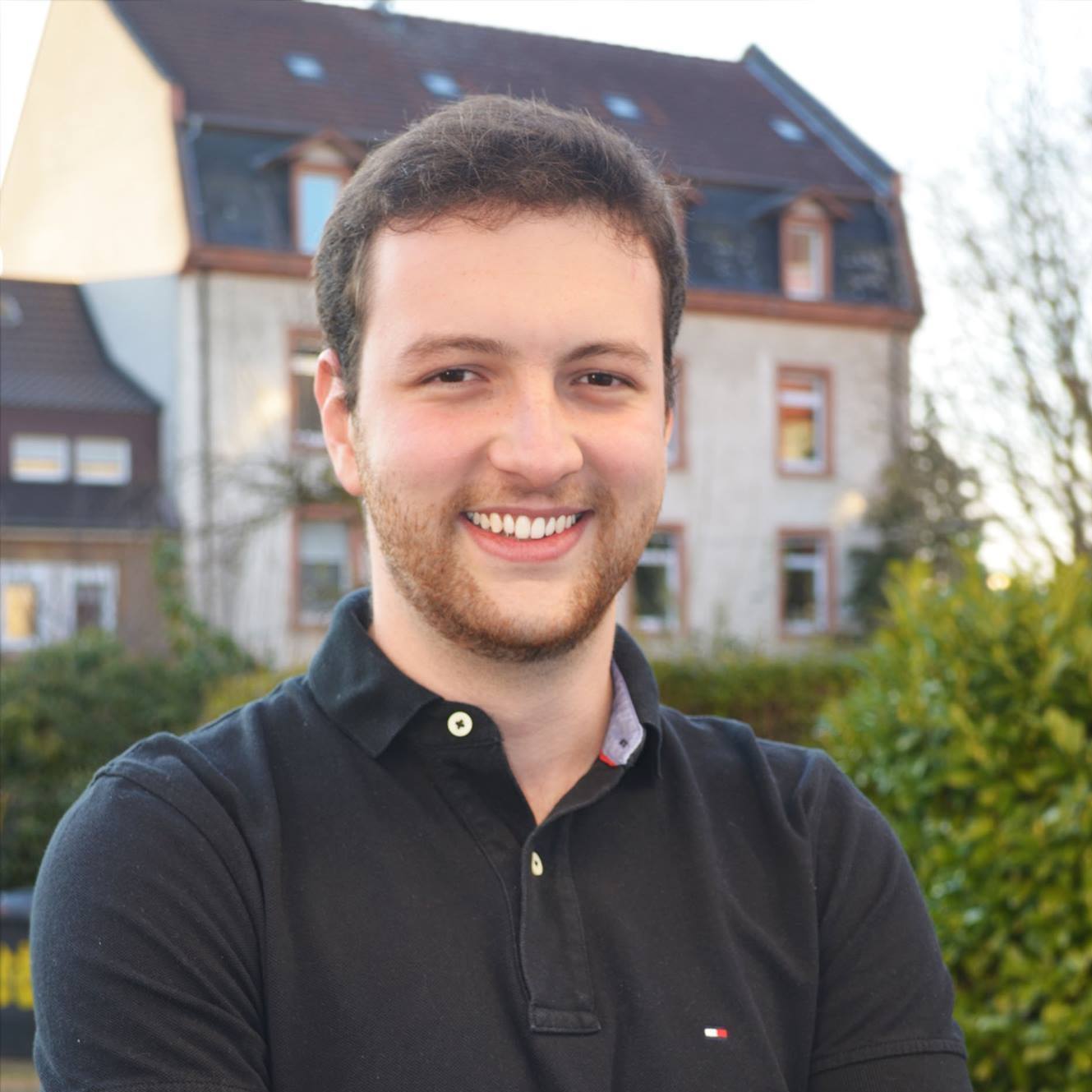}}]{André O. Françani} received the BSc degree in Electrical Engineering, with emphasis on Electronics, from the Polytechnic School of the University of São Paulo (EPUSP) in 2020. He also obtained his MSc degree in Electrical Engineering and Information Technology from the Technische Universität Darmstadt (TUD), Germany, as part of a double-degree program. Currently, he is pursuing a PhD in Electronic and Computer Engineering at the Aeronautics Institute of Technology (ITA). His main areas of interest are image processing, deep learning, and computer vision. His current research focuses on monocular visual odometry using deep learning algorithms.
\end{IEEEbiography}

\begin{IEEEbiography}[{\includegraphics[width=1in,height=1.25in,clip,keepaspectratio]{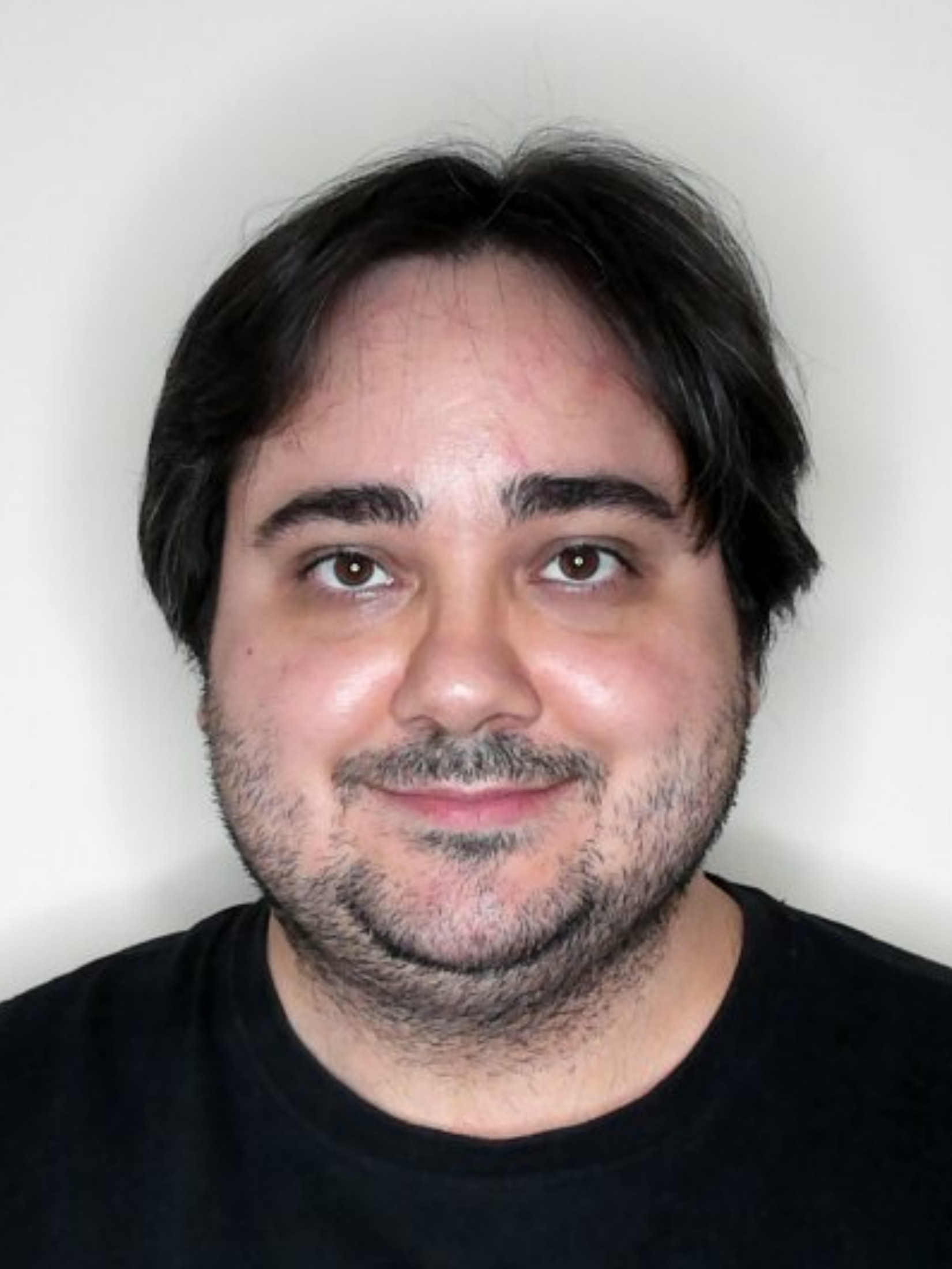}}]{Marcos R. O. A. Maximo} received the BSc degree in Computer Engineering (with Summa cum Laude honours) and the MSc and PhD degrees in Electronic and Computer Engineering from Aeronautics Institute of Technology (ITA), Brazil, in 2012, 2015 and 2017, respectively. Maximo is currently a Professor at ITA, where he is a member of the Autonomous Computational Systems Lab (LAB-SCA) and leads the robotics competition team ITAndroids. He is especially interested in humanoid robotics. His research interests also include mobile robotics, dynamical systems control, and artificial intelligence.
\end{IEEEbiography}

\end{document}